\documentclass[letterpaper, 10 pt, journal, twoside]{IEEEtran}

\IEEEoverridecommandlockouts                              % This command is only needed if
\makeatletter
\let\NAT@parse\undefined
\makeatother
\usepackage[numbers]{natbib}

\usepackage[pdftex]{graphicx}
\usepackage{amsmath}
\usepackage{amssymb}  % assumes amsmath package installed
\usepackage{subfig}
\usepackage{multirow}
\usepackage{array,booktabs}
\usepackage{diagbox}
\usepackage{balance}
\usepackage[ruled]{algorithm}
\usepackage{algpseudocode}
 \usepackage[final]{hyperref}
 \hypersetup{
     colorlinks=true,
     linkcolor=blue,
     filecolor=magenta,
     urlcolor=blue,
     citecolor=black
 }
\usepackage[font=small]{caption}
\usepackage{./rpm_packages/rpm_SIunits}
\usepackage{./rpm_packages/rpm_acronyms}
\usepackage{./rpm_packages/rpm_math}
\usepackage{./rpm_packages/rpm_misc}

\usepackage{soul,color}
\usepackage{lipsum}

\usepackage{xcolor}

\newcommand{\revision}[1]{{\textcolor{black}{#1}}}

\usepackage{tikz} % \checkmark

\title{LodeStar: Maritime Radar Descriptor for\\ Semi-Direct Radar Odometry}

\author{Hyesu Jang${}^{1}$, Minwoo Jung${}^{1}$, Myung-Hwan Jeon${}^{2}$, and Ayoung Kim${}^{1*}$

\thanks{Manuscript received: September, 19, 2023; November, 17, 2023; December, 25, 2023.}%Use only for final RAL version
\thanks{This paper was recommended for publication by Editor Giuseppe Loianno upon evaluation of the Associate Editor and Reviewers' comments.
This work was supported by the MOTIE (20024355), Korea.}
\thanks{$^{1}$H. Jang, M. Jung, and A. Kim are with the Department of Mechanical Engineering, SNU, Seoul, S. Korea {\tt\small [dortz, moonshot, ayoungk]@snu.ac.kr}}%
\thanks{$^{2}$M. Jeon is with the Institute of Advanced Machines and Design, SNU, Seoul, S. Korea {\tt\footnotesize myunghwan.jeon@snu.ac.kr}}%
\thanks{Digital Object Identifier (DOI): see top of this page.}
}

\begin{document}

%\onecolumn
\maketitle

\begin{abstract}
Maritime radars are prevalently adopted to capture the vessel's omnidirectional data as imagery. Nevertheless, inherent challenges persist with marine radars, including limited frequency, suboptimal resolution, and indeterminate detections. Additionally, the scarcity of discernible landmarks in the vast marine expanses remains a challenge, resulting in consecutive scenes that often lack matching feature points. In this context, we introduce a resilient maritime radar scan representation \textit{LodeStar}, and an enhanced feature extraction technique tailored for marine radar applications. Moreover, we embark on estimating marine radar odometry utilizing a semi-direct approach. \textit{LodeStar}-based approach markedly attenuates the errors in odometry estimation, and our assertion is corroborated through meticulous experimental validation.
The code will be available from \textbf{https://github.com/hyesu-jang/LodeStar}.

\end{abstract}
\begin{IEEEkeywords}
Range Sensing, Marine Robotics, SLAM
\end{IEEEkeywords}
\section{Introduction}
\label{sec:intro}

% 1. Background & Why Maritime radar odometry is needed?
%The maritime environment has perpetually presented an imminent risk, that human endeavors are yet to fully comprehend. Marine robotics are relentlessly striving to procure essential data while minimizing exposure to hazards in this formidable environment.
%Despite the endeavors, challenges persist in maritime sensing and decision-making. Typically, the perceivable object and the vessel are separated by considerable distances, rendering short-range sensors, such as cameras and \ac{LiDAR}, insufficient to generate pertinent features for the operational algorithm. Long-range sensors, such as Sonar or Radar, are standard practice in maritime operations, yet these are infamous for their subpar resolution.
\IEEEPARstart{W}{e} are situated within an epoch dominated by autonomous vehicles, where the sensors and algorithms underlying autonomous navigation have experienced exponential advancements.
In this context, an escalating demand for autonomous navigation in \ac{USV} seems logical. However, challenges persist in maritime sensing and decision-making due to environmental constraints. Typically, the perceivable object and the vessel are separated by considerable distances, thereby rendering short-range sensors, such as cameras and \ac{LiDAR}, insufficient in generating pertinent features for the operational algorithm. The use of long-range sensors, such as Sonar or Radar, is a standard practice in maritime operations, yet these sensors are infamous for their subpar resolution.
%Furthermore, the erratic motion of the vehicle due to wave-induced fluctuations hampers the consistency of sensor tracking.
%Therefore, the aforementioned factors impose significant constraints on the utilization of sensors that are commonly employed for autonomous navigation in the demanding maritime environment.

In the face of these challenges, substantial efforts continue to address environmental constraints that impede sensor performance. Of particular note is the progress in radar analysis, especially in adverse weather conditions where cameras and LiDAR systems are rendered inoperative. Despite the low resolution, enhanced processing of radar data could potentially enable place recognition~\cite{kim2020mulran, jang2023raplace} and provide LiDAR-equivalent vehicular odometry~\cite{adolfsson2021cfear, adolfsson2023tbv}. Given that radar imaging is a prevalent strategy in marine robotics, we have undertaken an effort to extract vehicular motion data from maritime radar in a manner analogous to the advancements witnessed in ground radar development.

%FIGURE
\begin{figure}[!t]
    \centering
    \vspace{2mm}
    \includegraphics[width=\columnwidth]{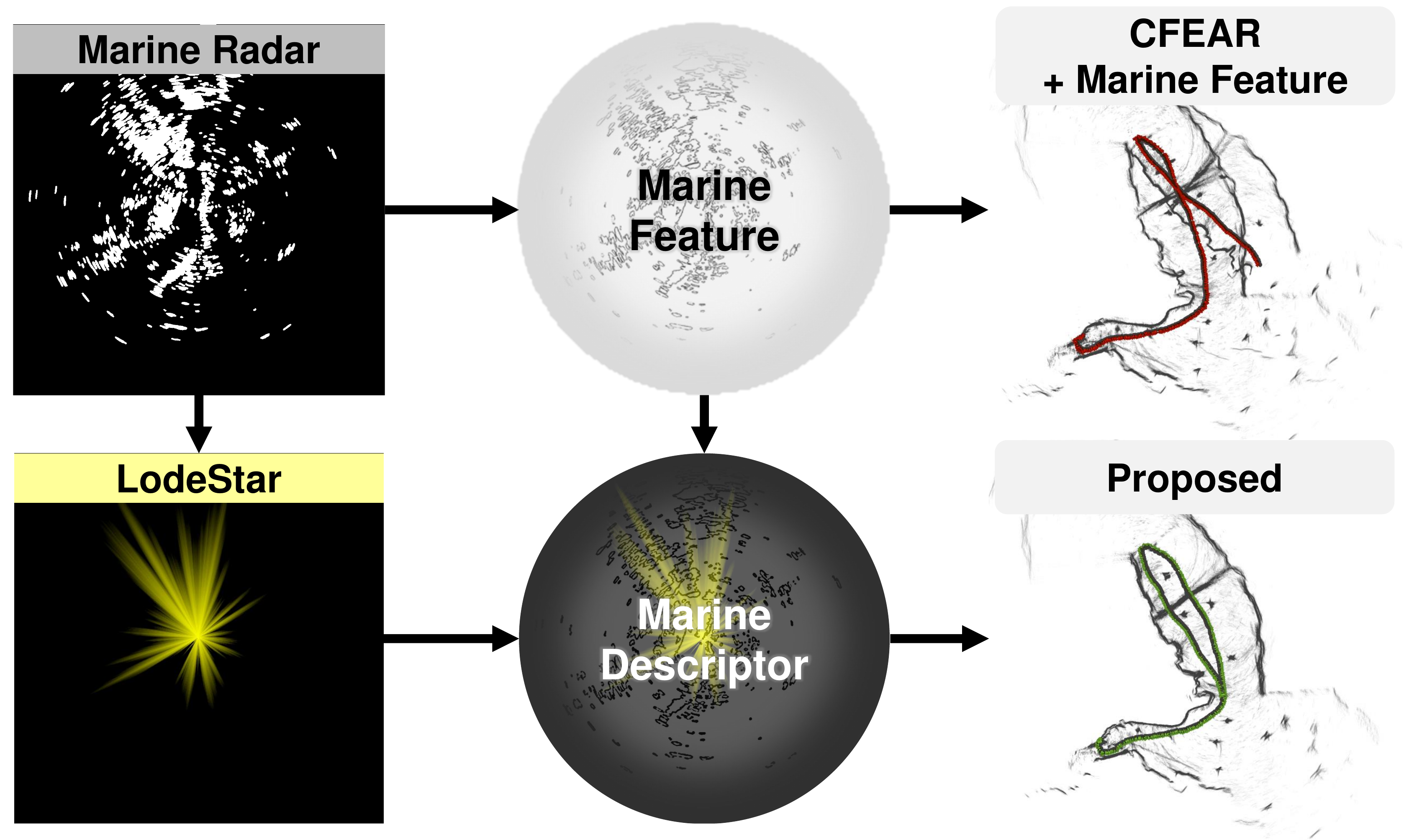}
    \caption{Terrestrial radar-based odometry estimation techniques cannot be directly transposed to marine radar systems. As illustrated in the upper right side, even the most advanced radar odometry methodologies fail to generate accurate trajectories. Nonetheless, our maritime place descriptor LodeStar, incorporated with marine-specific features, effectively captures the rotational dynamics of subsequent frames, leading to enhanced odometry correction.}
    \label{fig:intro}
    \vspace{-6mm}
\end{figure}
%FIGURE

% 2. What is the difference between ground and Maritime radar? & What are the challenges for maritime odometry?
The effective use of marine radar necessitates overcoming several key challenges. %Firstly, the discrete and limited variety of \ac{RCS} levels require mitigation. While ground radar generates a broad range of \ac{RCS} values, marine radar, with its focus on longer-range detection, restricts the receipt of a diverse range of intensity detections.
Firstly, \ac{RCS}, intensity return in radar measurement, is quantized and nearly binary in marine radar. While ground radar generates a broad range of \ac{RCS} values, marine radar restricts the receipt of a diverse range of intensity detections.
%Secondly, the coastal contour must be accurately represented. Whereas buildings and roads serve as distinctive feature points for ground radar, in oceanic environments, the most reliable features predominantly lie within the contours of the coast.
Second challenge is the high reliance on coastal contour. Whereas buildings and roads serve as distinctive features for ground radar, in oceanic environments, the most reliable features predominantly lie within the contours of the coast. Accurate representation of coastal places is required for marine radar.
%Lastly, attention must be devoted to reducing ambiguity and uncertainty in matching due to the significant size of the pixel units. Given that the radar's detection range spans about a 2-3 kilometer, the unit pixel of the Cartesian imagery results inherently starts from the meter unit. Consequently, an error of a single pixel can lead to significant drift in marine radar odometry.
Last one is the ambiguity and uncertainty of detected radar pixels. A false alarm of a single pixel can lead to significant drift in marine radar odometry, subsequently leading to substantial errors in the ensuing \ac{SLAM} operation. Given that the radar's detection range spans about a $2-3\km$, the unit pixel of the Cartesian imagery results starts from the meter unit.

% 3. What we have done?
In this paper, we propose a methodology for maritime radar-only odometry estimation that addresses the aforementioned challenges. We design a counterstrategy to the limited \ac{RCS} levels that extract marine radar feature points and contours. Additionally, employing our dense radar descriptor, \textit{LodeStar}, we estimate the vehicle rotation and generate a rotated pointcloud to enable swift and accurate convergence of sparse point matching.
\figref{fig:intro} provides an visual representation of procedures and improvements.
The detailed contributions of our paper are as follows:

\begin{itemize}
    \item \textbf{LodeStar: Maritime Radar Descriptor}\\
    The \textit{LodeStar} is a star-shaped descriptor encapsulating the radial context information derived from maritime radar. The periodic form of the descriptor enables the conduct of circular cross-correlation, effectively utilized for dense searching of optimal rotational change values.
    
    \item \textbf{Robust Maritime Features for Radar}\\
    Identifying plausible feature points within marine radar data poses a complex problem due to the inherent ambiguity of radar detection results. Therefore, we have defined specific feature points within the marine radar data that assist in deriving point normal.

    \item \textbf{Semi-direct Maritime Odometry Estimation}\\
    While dense methods offer comprehensive search capabilities, they suffer from \revision{high computation time complexity.} Conversely, sparse methods, while efficient, struggle to fully grasp the overall context of sequential inputs. By amalgamating the strengths of both methods, we propose a combined approach that has the potential to significantly enhance the accuracy of odometry.
\end{itemize}

\section{related work}
\label{sec:relatedwork}
% Scanning!!!!!

%In this section, we present an overview of radar odometry methodologies, primarily focusing on terrestrial radar odometry owing to its relevance to the subject matter of our research. Numerous studies have been conducted in this field, necessitating a targeted examination of the subset most relevant to our work. 
The applicability of scanning radar used terrestrially is demonstrated by its similarity to marine radar. Following the taxonomy by \cite{abu2023radar}, we categorize the existing work into three main strands: sparse, dense, and hybrid. In addition, we further cover the studies on marine radar odometry.

%====================================================================================%
\subsection{Sparse Radar Odometry}
Sparse radar odometry encompasses methods based on features and scan-matching. HERO~\cite{burnett_rss21}, a feature-based method, is notable for its use of unsupervised learning to extract features from Cartesian radar images. 
%Leveraging these features, it employs a non-differentiable classic estimator for probabilistic inference in trajectory estimation. 
The research conducted by \citeauthor{lim2023orora}~\cite{lim2023orora} adopts the mechanism of re-estimating outliers during the process of rotation and translation estimation.
The work of \citeauthor{cen2019radar} \cite{cen2019radar, cen2018precise} makes use of non-visual features, derived from the statistical properties of the radar power return. %The initial research \cite{cen2018precise} involves employing Singular Value Decomposition (SVD) to estimate the relative motion between two scans. This methodology was subsequently enhanced with the improved feature extraction algorithm in \cite{cen2019radar}.

Contrasting the feature-based methods, scan matching techniques do not require correspondences between two scans. A notable example of this approach is CFEAR \cite{adolfsson2021cfear, adolfsson2022lidar}, which determines a transformation that minimizes the point-to-line distance between a scan and a keyframe. The extraction of normal vectors in this method is conducted in two stages: initially, the top $k$ points with the strongest returns are retained and subsequently clustered to form a surface. Based on these surfaces, the normal vectors are computed using the eigenvectors derived from the covariance matrix of each surface. Furthermore, \citeauthor{kung2021normal}~\cite{kung2021normal} introduces a probabilistic radar submap, constructed based on sparse Gaussian Mixture Models. They then employ the \ac{NDT} to compute the transformation between two consecutive scans.

In odometry techniques that employ sparse methodologies, the availability of robust feature points is directly linked to the performance. Nevertheless, procuring such robust features and assigning their correspondences present considerable challenges in marine environments.

%-----------------------------------------------------------------%
\subsection{Dense Radar Odometry}
Dense radar odometry methods utilize complete radar scans as input for the calculation of relative transformations between scans. One of the early approaches \cite{checchin2010radar} employs the \ac{FMT}, an image restoration method, to calculate the relative transformations. %This technique involves transforming the polar image into the frequency domain using the \ac{FFT}. Cross-correlation then facilitates the estimation of rotations between images. After rotation correction, both \ac{FFT} and cross-correlation are again utilized to compute the translation.
Similarly, \citeauthor{park2020pharao} \cite{park2020pharao} applied \ac{FMT} in two stages. Initially, rotation and initial estimate of translation are determined from a downsampled image. Subsequently, the translation part is recalculated using a full-resolution image. By combining this methodology with keyframe selection and graph optimization, it can achieve heightened accuracy.

In contrast to model-based odometry, research has also been conducted on dense radar odometry employing deep learning methods. For instance, \citeauthor{barnes2019masking} \cite{barnes2019masking} \revision{utilized} a U-net style \ac{CNN} to create masks that suppress image noise. %The filtered images are then cross-correlated to identify the maximum cross-correlation, which is then used to estimate the relative transformation. 
A subsequent improvement \cite{weston2022fast} decouples the rotation and translation. Leveraging the translation invariance property of the Fourier Transform in polar coordinates, the search time for identifying the maximum cross-correlation can be significantly reduced.

Direct methods exhibit robustness in analyzing the context of the scene, facilitating an approximate estimation of the vessel's pose. Nonetheless, due to the low resolution of marine radar imagery, achieving precise pose estimation remains a formidable challenge in \revision{direct methods.}
%-----------------------------------------------------------------%
\subsection{Hybrid and Marine Radar Odometry}

Given the unique strengths of both sparse and dense odometry, their combined use could yield enhanced benefits. \citeauthor{8995552} \cite{8995552} employed the sparse method for translation estimation and the dense method for rotation computation. However, this strategy does not optimally leverage the concurrent benefits of sparse and dense methodologies, choosing instead to apply them independently.
%Specifically, in marine environments where the measurement data may be limited, the sparse methodology encounters certain limitations. Conversely, the dense methodology could result in local minima solutions in the presence of considerable environmental noise. 
%Although alternative sensor-based marine odometry methods exist \cite{terzakis2017monocular}, radar-based techniques hold considerable advantages due to their capacity to detect distant objects. 
Recognizing this, the potential benefits of a hybrid approach to radar odometry, harnessing the strengths of both sparse and dense methods, are worth investigating. However, existing marine odometry with scanning radar have primarily focused on features \cite{5509262,han2019coastal, schiller2022improving}. \revision{To bridge this gap, our approach introduces semi-direct maritime odometry, integrating the dense descriptor \textit{LodeStar} with sparse maritime feature extraction and matching.} %This is an innovative approach that promises to expand the capabilities of marine odometry.

%====================================================================================%

\section{Method}
\label{sec:method}
\begin{figure}[!t]
    \centering
    \includegraphics[width=0.85\columnwidth]{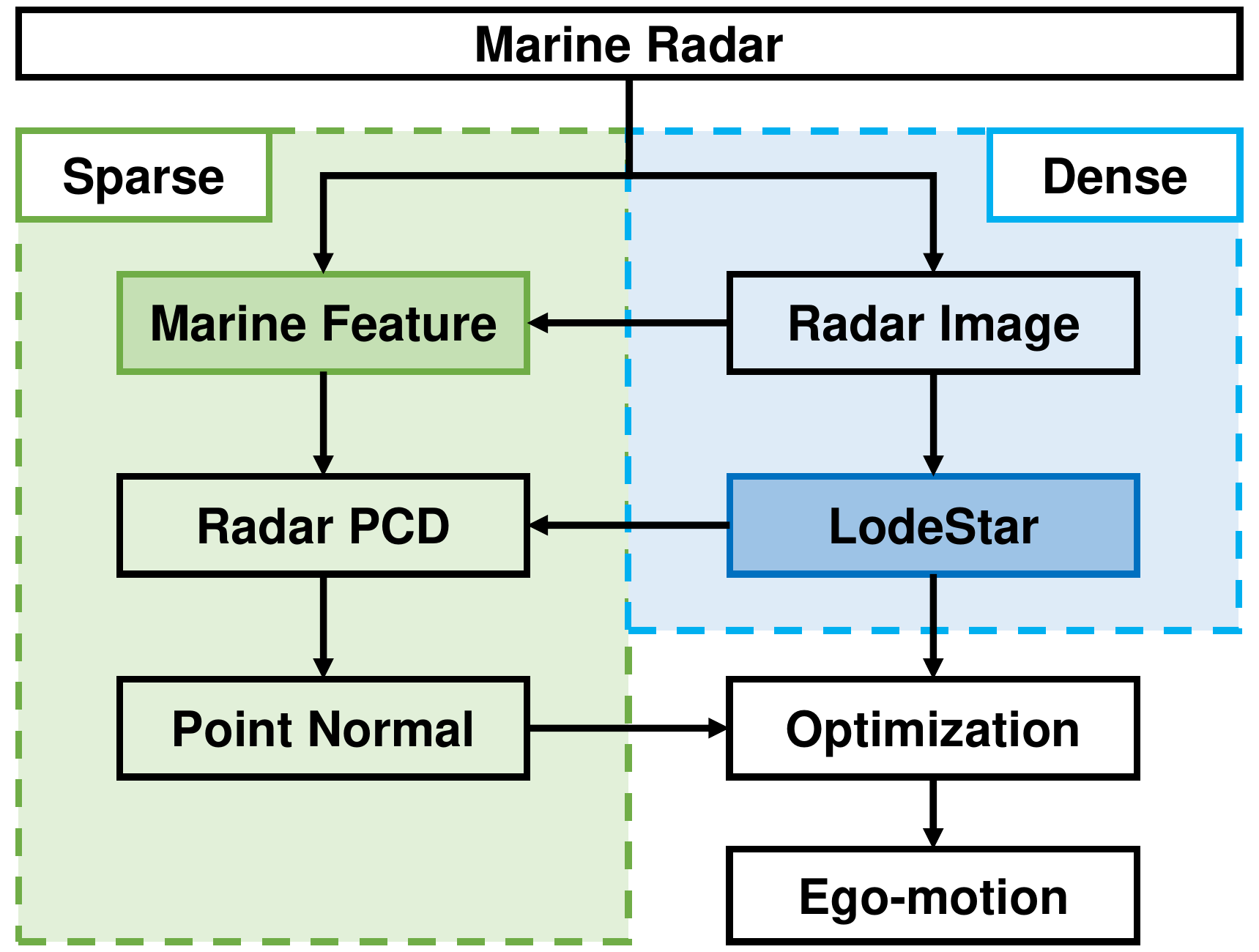}
    \caption{\revision{The proposed framework for maritime odometry estimation.} Exclusively using marine radar imagery, we extract pertinent marine features and the LodeStar descriptor. Subsequently, we construct an initial rotated point cloud and identify correspondences between two consecutive frames via a point normal-based approach. This process derives a rotation-enhanced ego-motion estimation.}
    \label{fig:overview}
    \vspace{-6mm}
\end{figure}

The method proposed in this paper is an odometry estimation, which is exclusively reliant on maritime radar data. As illustrated in \figref{fig:overview}, the only measurement input used is derived from maritime radar.
Our methodology uses two branches: we integrate radar image data into the matching process and utilize point cloud data for feature-driven matching.
Maritime radar data is presented in the form of Cartesian image data. This data is leveraged to deduce coarse rotational information using a robust descriptor called as \textit{LodeStar}. To enhance computational efficiency during optimization and to boost overall accuracy, we provide \textit{LodeStar}-based rotation-corrected images to carry out feature-based transformation estimation.
We transform these rotationally corrected image data into point cloud data. In order to focus solely on the most dependable point clouds, we extract feature points specific to maritime contexts and compute the vehicle motion using a point-to-normal matching method.
Further details and the comprehensive process will be discussed and illustrated in the subsequent sections of this paper.

%FIGURE

\subsection{LodeStar Descriptor}
\begin{figure}[!t]
    \centering
   % \vspace{-1mm}
    \includegraphics[width=0.98\columnwidth]{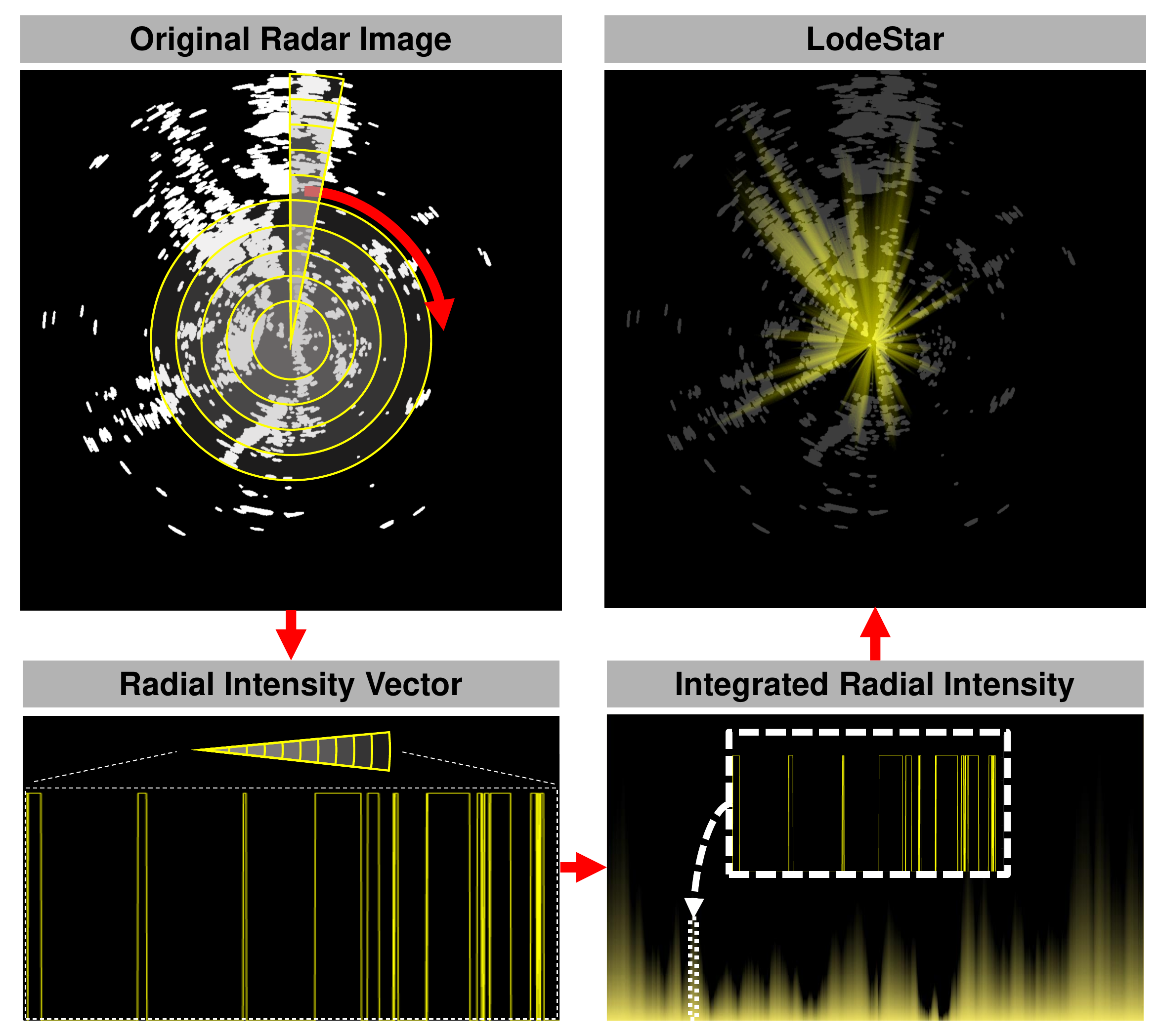}
    \vspace{-1mm}
    \caption{Details of the LodeStar descriptor. From the primary radar imagery, we compute a radial intensity vector for every $\theta$. Each of these vectors is subsequently integrated to constitute a column within the integrated radial intensity matrix. By iterating this process over a period of 2$\pi$, we synthesize the final descriptor.}
    \label{fig:lodestar}
    \vspace{-5mm}
\end{figure}

%Maritime radar data exhibits a distinctive trait: the areas covering harbor-side territories can be approximated, albeit with inherent data sparsity. To exploit this feature, previous methodologies \cite{han2019coastal} attempted to utilize harbor contour information. However, these approaches underutilized the harbor territories, primarily because these areas are sporadically populated with variable intensity values.
%The unpredictability of these values, which change with each frame, makes the pixels within the harbor challenging for feature point utilization. 
In our effort to design a robust descriptor, we opted to encapsulate the spatial information of the harbor area, thereby addressing the challenges of prior methodologies.
The foundational idea of \textit{LodeStar} is rooted in the Radon transform \cite{jang2023raplace}\revision{; instead} our approach simplifies by focusing on radial-wise integration.  A detailed visual depiction for \textit{LodeStar} is presented in \figref{fig:lodestar}. 

\revision{The descriptor $L(\theta)$ corresponding to azimuth angle $\theta$ is derived} as per
\begin{equation}
\centering
\begin{split}
L(\theta) &= \int_{0}^{r_{\max}} I(x,y) \,dr \\
& = \int_{0}^{r_{\max}} I(r_{\max}-r\sin{\theta},r_{\max}+r\cos{\theta}) \,dr \\
& = L(\theta + 2\pi n), \quad [n \in \mathbb Z].
\end{split}
\label{eq:lodestar}
\end{equation}
The radar's maximum detection range is represented as $r_{max}$, while the unit range, denoted as $r$, varies from the center of the radar image $I(x,y)$. By avoiding integration across the entire image spectrum, we are able to curtail the computational burden, thus enhancing efficiency.
Given the need to infer rotational changes from the descriptor, we extended $L(\theta)$ as a periodic function. Cross-correlation was employed to compute the disparity between the two descriptors. The periodic nature of $L(\theta)$ made it possible to acquire a circular correlation result.
\begin{equation}
\centering
\theta_{L_i} = \underset{\theta}{\arg\max}\left(\int_{0}^{2\pi} L_{i}(\theta+\mu) L_{i+1}(\mu) \,d\mu \right)
\label{eq:ccc}
\end{equation}
The circular cross-correlation outcome of the LodeStar descriptor enables us to define the rotation difference between frames $i$ and $i+1$ as $\theta_{L_i}$. This rotation value plays a pivotal role in generating angular-corrected point cloud data.
\begin{equation}
\centering
\begin{bmatrix} x' \\ y' \end{bmatrix} = \begin{bmatrix}\cos(\theta_{L_i}) & -\sin(\theta_{L_i}) \\
\sin(\theta_{L_i}) & \cos(\theta_{L_i})
\end{bmatrix} \begin{bmatrix} x \\ y \end{bmatrix}
\label{eq:rotimg}
\end{equation}

%FIGURE

\subsection{Maritime Radar Feature Extraction}
We can transmute radar imagery into a set of pointcloud data, utilizing two-dimensional bird-eye view coordinates and the inherent pixel intensity values. Nevertheless, defining cogent features within the complex, fluctuating maritime environment remains an intricate task. This section elucidates the methodology for extracting distinct features from marine-based radar datasets.

\subsubsection{Contour Extraction}
Contour extraction and matching offers a macro-environment perspective without relying on internal data. Excluding points within the contour is beneficial for point normal estimation, given that radar data cannot comprehensively populate the interior of the contour. Additionally, the contour presents dependable candidates for the $k$-nearest feature extraction, mitigating the inclusion of noise. The polygonal contour can be computed using the polar image and the B-spline method as introduced by \citet{han2019coastal}, but in our approach, we opted for simple contour extraction from the \ac{RCS}. \revision{We experimentally noticed that some radar products exhibit the lowest \ac{RCS} values at contours, while others depict the highest. Based on the premise that the reflectivity at the boundary differs from that of the planar area, we implemented a high-pass or low-pass filter for contour extraction depending on the radar type.}

\subsubsection{$k$-nearest Feature Candidates}
On-ground scanning radars present a diverse and continuous range of \ac{RCS} levels, and common practice is to utilize $k$-strongest points as potential candidates as \revision{done by} \citet{adolfsson2021cfear}. Unlike ground radar, maritime radar exhibits discrete RCS levels, with a significant number of points sharing identical RCS values. To pinpoint the candidates for point normal extraction, we employ a $k$-nearest feature extraction approach instead of $k$-strongest.
Adopting the $k$-strongest points in the extant methodology was for selecting robust and dependable points. However, given the almost binary nature of the RCS of marine radar, extracting the most prominent point is ineffective. Furthermore, marine radar is a kilometer-scale long-range sensor, \revision{thus} the uncertainty associated with distant points is substantial. We select points nearest to the center from every angle to mitigate this uncertainty.
In conclusion, we perceive feature points as the non-overlapping, nearest contour points.

\subsubsection{Overlapping Data Elimination}
\begin{figure}[!t]
    \centering
   % \vspace{-1mm}
    \subfloat[Overlap of two images\label{fig:feature_overlap1}]{
		\includegraphics[height=0.48\columnwidth]{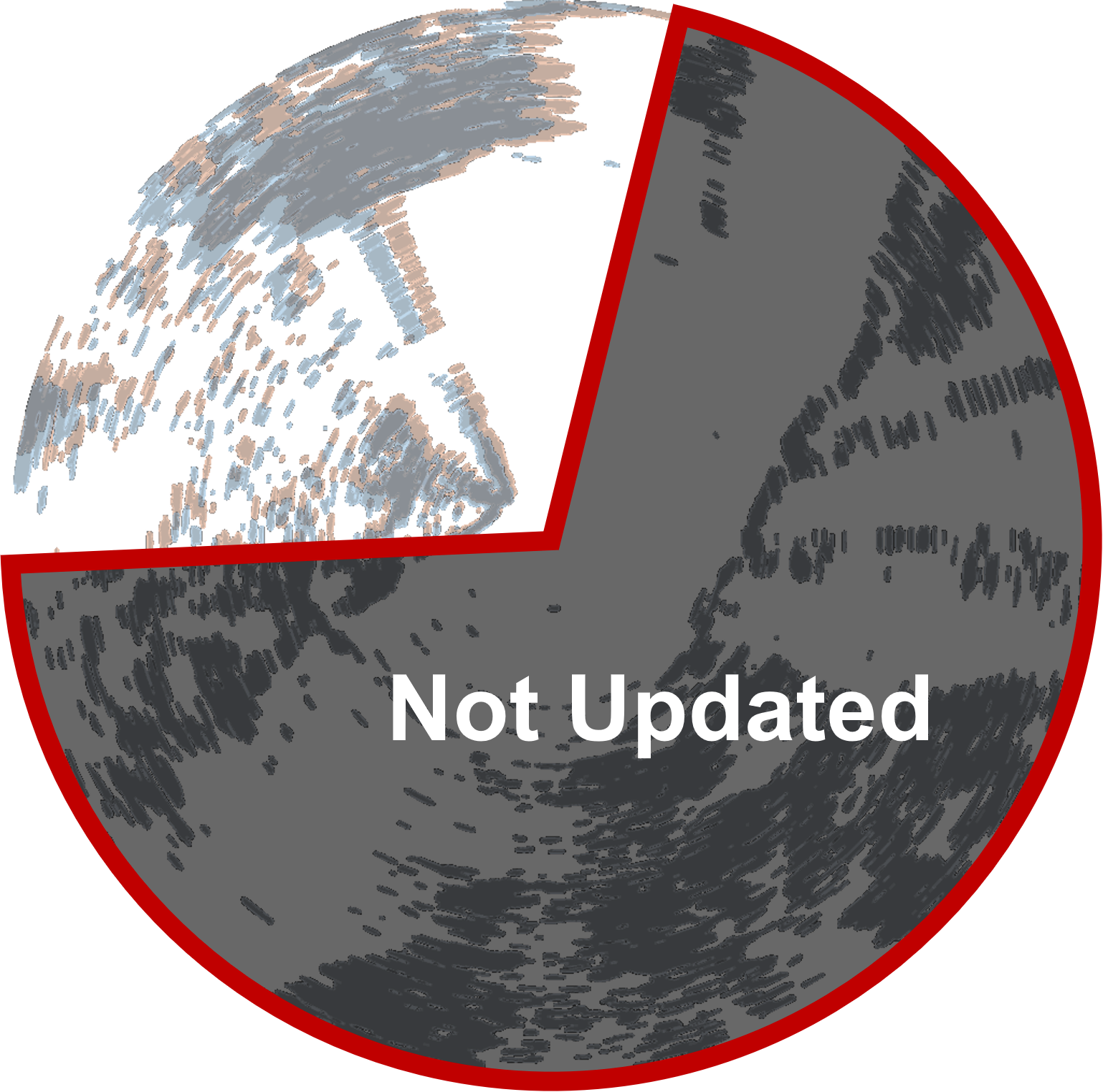}
    }
    \subfloat[Revolution scan delay\label{fig:feature_overlap2}]{
		\includegraphics[height=0.48\columnwidth]{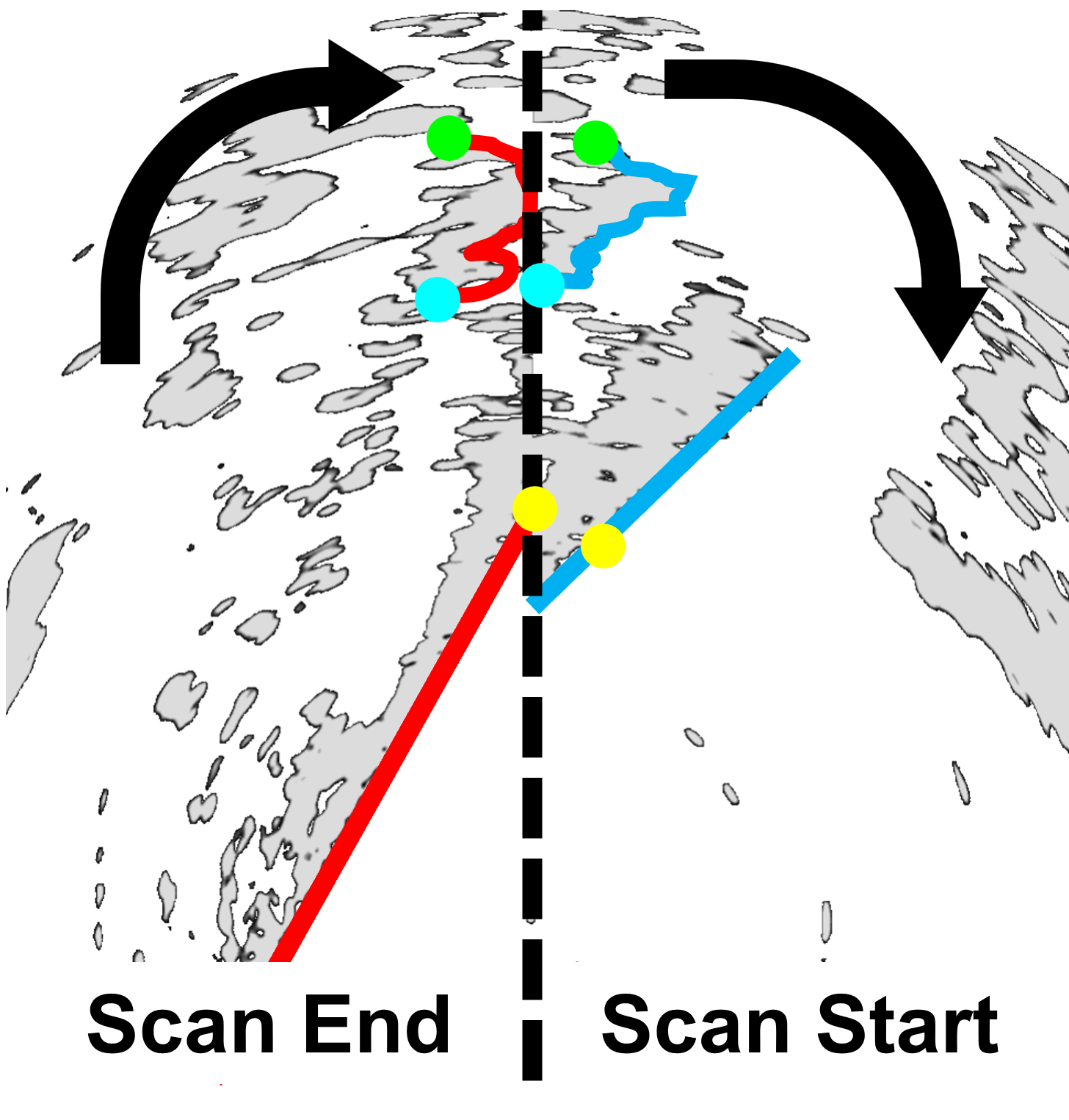}
	}\\
    \subfloat[Contour and nearest point extraction\label{fig:feature_contour}]{
		\includegraphics[width=0.98\columnwidth]{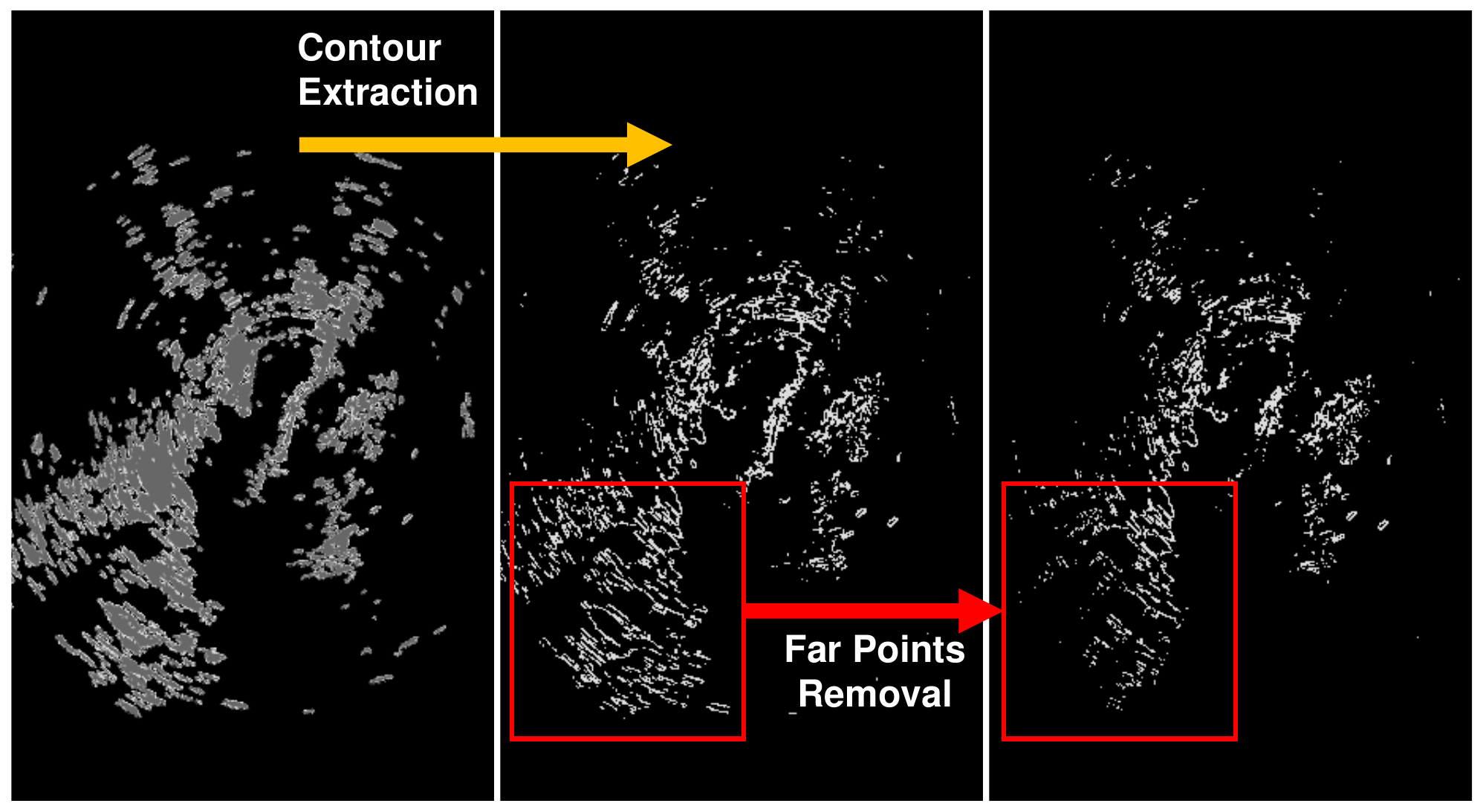}
	}
    \vspace{-1mm}
    \caption{(a) In our dataset, certain regions remain un-updated due to the low update rate. An accumulation in these overlapping areas could lead to unexpected convergence behaviors. (b) Given that radar imagery inherently constitutes scan-rotated images, there exists a discrepancy between the commencement and conclusion of the scans, primarily attributed to temporal lags. This discrepancy is especially pronounced during sharp rotational movements. (c) Initially, we delineate the contour, subsequently identifying proximal points to ensure consideration extends to the wide area.}
    \label{fig:feature}
    \vspace{-5mm}
\end{figure}

Radar imagery is derived from the systematic, rotational alignment of ongoing radar scans, updated dynamically by scan vector $V_t(t)$ published at time $t$. This indicates that the publication of a radar image $V_{t}(t_{i}:t_{i+1})$ does not necessitate a full revolution completion of the radar scan. 
Should the radar image publication frequency be higher than the radar's full-rotation frequency, consecutive image frames may contain identical data, as demonstrated in \figref{fig:feature}. Conversely, rotational and translational glitch may arise if the radar image data is only refreshed following the scanning of all angular measurements, as illustrated in \figref{fig:feature}. This predicament calls for precise regulation of the radar image frequency.
Pursuing optimal radar imagery necessitates a balance between a high scan update frequency and preventing radar image overlap.
\begin{equation}
\centering
\small
\begin{split}
I_{i}(x,y) & = V_{t}(t_0:t_3)\\
& = V_{t}(t_0:t_1) + V_{t}(t_1:t_2) + V_{t}(t_2:t_3)\\
&\simeq V_{t_1}(t_0:t_1) + V_{t_2}(t_1:t_2) + V_{t_3}(t_2:t_3) = I_{i}^{part}\\
&\simeq V_{t_3}(t_0:t_3)  = I_{i}^{full}\\
\end{split}
\label{eq:overlap}
\end{equation}
To encompass distinct radar scan sets within a single image, we escalated the frequency of radar image data and implemented a subscription dropout to eradicate overlapping regions $I_{i}^{part}(x,y)$. If the radar images are derived from a full-rotation scan, we exploit the integrated image $I_{i}^{full}(x,y)$ at the last scan time. We portrayed the implications of overlap and anomalies for individual datasets in the results section.
%FIGURE

\subsection{Point Normal Matching and Optimization}
In our detailed motion estimation framework, we employed the point normal matching method established by \citet{adolfsson2021cfear} which represents a current benchmark in the field of radar odometry estimation. However, the method is inherently predisposed toward using scanning radars within the context of autonomous vehicles. Consequently, its direct application to marine radar imagery yields suboptimal results due to the scarcity of reliable candidates. Hence, it is necessary to recalibrate its configuration to better adapt to marine radar environments. This adjustment process has been thoroughly addressed in the preceding sections of this work.
We have made significant strides in minimizing rotational drift and the degree of uncertainty by integrating a bespoke descriptor. Moreover, instead of defaulting to the top-k strongest radar points as inputs, we have elected to use feature points that were derived from the previous section.
%rewrite from here
With the uncertainty-minimized data, we derived $n$ number of surface point and normal pairs $\upsilon(n) = (p(n),\eta(n))$. Pointwise scan registration is conducted by minimizing the point-to-point error function. For $a$ and $b$ satisfying the condition $[\upsilon_i(a) \in \upsilon_i,\upsilon_{i+1}(b) \in \upsilon_{i+1}]$, we defined error function as \eqref{eq:error}.
\begin{equation}
\centering
\small
\epsilon(\upsilon_i(a), \upsilon_{i+1}(b),\mathbf{T}_i^{i+1})  = \|p_{i+1}(b) - (R_i^{i+1}p_i(a) + \tau_i^{i+1})\|^2
\label{eq:error}
\end{equation}

To find the optimal transformation $\mathbf{T} = [ R , \tau ]$ with rotation $R$ and translation $\tau$, we adopted Cauchy loss function $\mathcal{L}(\epsilon) = \log(1+\epsilon^2)$ based argument. \revision{The Cauchy loss function exhibits a logarithmic escalation with residuals, thereby demonstrating enhanced resilience to outliers compared to alternative loss functions. This attribute is particularly suitable for radar point cloud processing.} \revision{To enhance the reliability of the model, we incorporated a similarity weight $\alpha$, calculated from the pair $v$ as described in \cite{adolfsson2021cfear}.}
\vspace{-0.5mm}
\begin{equation}
\centering
f(\upsilon_i, \upsilon_{i+1},\mathbf{T}_i^{i+1}) = \sum_{\forall {a,b}}\alpha\mathcal{L}(\epsilon) = \sum_{\forall {a,b}}\alpha\log(1+\epsilon^2)
\label{eq:loss}
\end{equation}
\vspace{-3mm}
\begin{equation}
\centering
[\Delta x,\Delta y,\theta_{P_i}] = \underset{x,y,\theta}{\arg\min} f(\upsilon_i, \upsilon_{i+1},\mathbf{T}_i^{i+1})
\label{eq:registration}
\end{equation}

Upon optimization, the transformation matrix was ascertained by minimizing the discrepancy between the two point cloud sets. The consequent alterations in rotation can be symbolized as the sum of two angles, $\Delta\theta_i = \theta_{L_i}+\theta_{P_i}$.

\section{experiment}
\label{sec:experiment}
\subsection{Evaluation Environment Configuration}
\subsubsection{Maritime Radar Datasets}
Given the unique environmental characteristics, there are few publicly available open datasets for maritime odometry estimation. In order to execute and validate our algorithm, we utilized two datasets, the Pohang Canal Dataset \cite{chung2023pohang} and our proprietary dataset (denoted as Ulsan). 
%Both dataset acquired data with \textit{SIMRAD} marine radar. 
The Pohang Canal Dataset offers 0.77Hz to 1Hz full Cartesian radar image data, devoid of overlapping regions. On the other hand, our dataset provides 1Hz full Cartesian radar image data, but with overlapping radar scans. \revision{The experimental procedure was executed in an offline environment; however, the feasibility of online implementation is assured owing to the low frequency of sensor data acquisition.} Detailed numerical characteristics of both datasets have been tabulated and can be found in \tabref{tab:dataset}.

\begin{figure}[!t]
    \centering
    \includegraphics[width=\columnwidth]{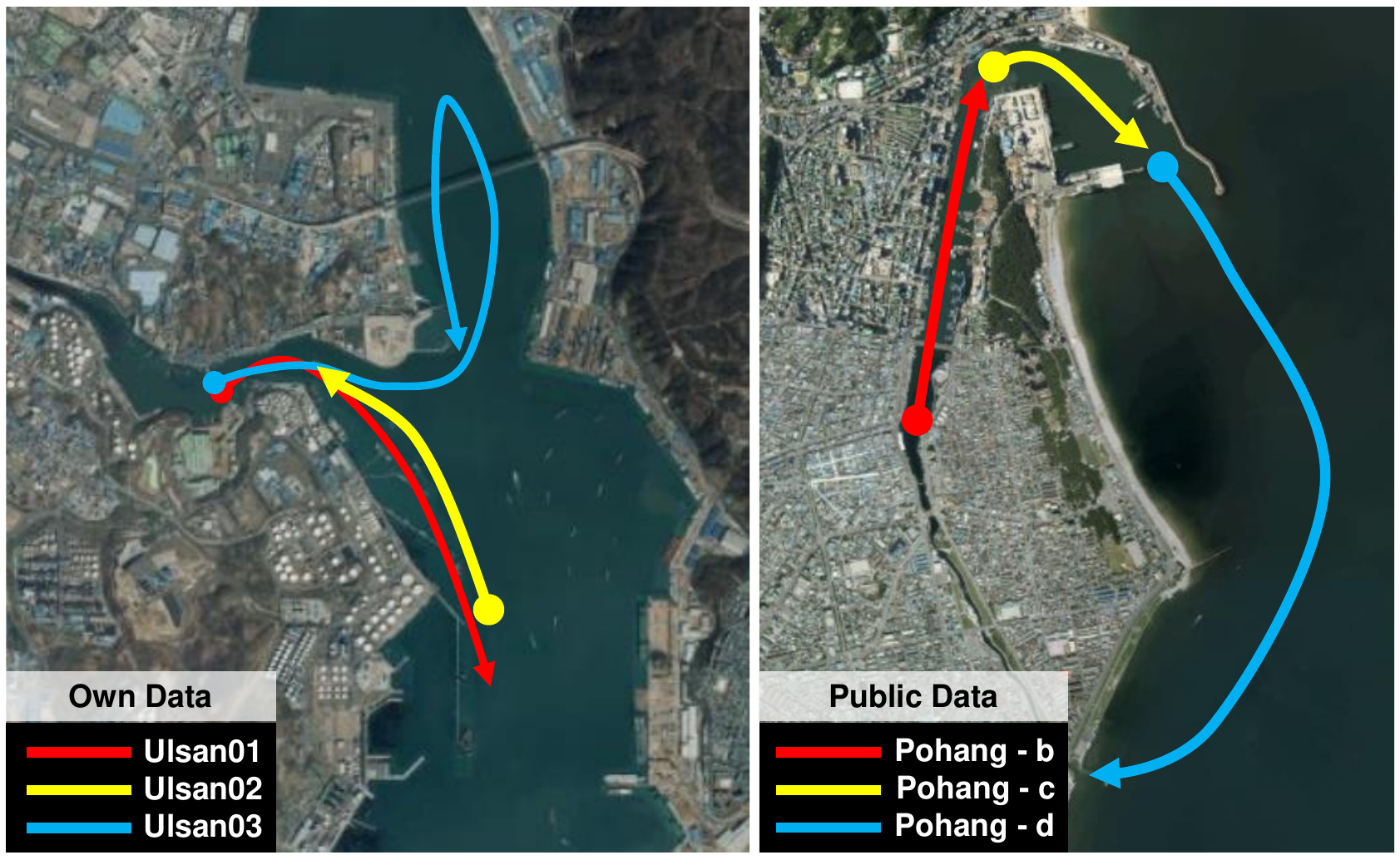}
    \caption{For the given dataset, a cursory classification of the routes can be made as follows: \texttt{Pohang-b,c} and \texttt{Ulsan01,02} represent more straightforward routes. In contrast, \texttt{Pohang-d} and \texttt{Ulsan03} pose greater challenges for odometry estimation.}
    \label{fig:dataset}
    \vspace{-2mm}
\end{figure}

\begin{table}[t!]
\resizebox{\linewidth}{!}{
\begin{tabular}{cc|cccccc}
\multicolumn{2}{c}{\textbf{Dataset}}                    & \textbf{Distance}  & \textbf{In-Port} &\textbf{Out-Port} &\textbf{Coastal} &\textbf{Shape}\\ \hline\hline
\multicolumn{1}{c}{\multirow{3}{*}{\texttt{Ulsan}}} & \texttt{01}    & 2.67km  & Y & Y & N & Curved           \\ 
\multicolumn{1}{c}{}                       & \texttt{02}   & 2.49km  & Y & Y & N & Linear             \\ 
\multicolumn{1}{c}{}                       & \texttt{03} & 5.08km    & Y & Y & N & Steep turn       \\  \hline
\multicolumn{1}{c}{\multirow{5}{*}{\texttt{Pohang}}} & \texttt{00-b}  & 1.62km   & Y & N & N & Linear       \\
\multicolumn{1}{c}{}                       & \texttt{01-b}  & 1.95km   & Y & N & N & Linear       \\
\multicolumn{1}{c}{}                       & \texttt{00-c}  & 0.76km   & N & Y & N & Curved       \\
\multicolumn{1}{c}{}                       & \texttt{00-d}  & 3.37km   & N & Y & Y & Steep turn       \\
%\multicolumn{1}{c}{}                       & 1-3  & HIHIHI   & N & Y & N & Curved      \\
\multicolumn{1}{c}{}                       & \texttt{01-d}  & 3.53km   & N & Y & Y & Steep turn  \\\hline
\end{tabular}}
\caption{Radar Dataset Attributes}\label{tab:dataset}
\vspace{-4mm}
\end{table}

We obtained the dataset in a port environment, with the marine radar and \ac{LiDAR} equipped vessel. In our dataset, radar imagery is distinguished by incorporating radar scans at a higher frequency than the Pohang canal dataset. This results in the production of superimposed radar images. As illustrated in \figref{fig:dataset}, the right panel portrays the rudimentary trajectories associated with our dataset. Specifically, \texttt{Ulsan01} and \texttt{Ulsan03} initiate from the inner port vicinity, proceeding towards the external port regions. \texttt{Ulsan01} encompasses a trajectory with a modest curvature, whereas \texttt{Ulsan03} manifests a steep turn, susceptible to causing disruptions in rotational tracking. Conversely, \texttt{Ulsan02} originates from the outer port domain, navigating linearly into the inner port.

The Pohang Canal Dataset~\cite{chung2023pohang} is delineated into four distinct regions, each assigned a unique numerical identifier: \textit{a}-Narrow canal area, \textit{b}-Inner port area, \textit{c}-Outer port area, and \textit{d}-Near-coastal area. Due to unstable radar data within the narrow canal region, segments labeled \texttt{Pohang-a} were excluded from the evaluation process. Additionally, in light of the absence of GPS data for \texttt{Pohang01-c}, this segment was not subjected to assessment. Consequently, our algorithm was scrutinized across five datasets, specifically \texttt{Pohang00-b}, \texttt{Pohang00-c}, \texttt{Pohang00-d}, \texttt{Pohang01-b}, and \texttt{Pohang01-d}. Pohang sequences facilitated observing the algorithm's performance across linear, curved, and near-coastal trajectories.

\subsubsection{Evaluation Criteria}
We compared our devised algorithm with contemporarily advanced methodologies that have been published recently. For conducting an in-depth comparison with \ac{LiDAR}-based odometry operating within a marine environment, we have opted for two distinct \ac{LiDAR} odometry estimation procedures~\cite{chen2022direct,vizzo2023kiss}. As for the radar odometry, we have incorporated representative techniques for both metric and learning-based sparse estimation~\cite{adolfsson2021cfear,burnett_rss21}, given the absence of a publicly available open-source resource for dense estimation.
In the main result \tabref{tab:result}, we delineate the influence of our marine-specific features; \textit{Contour} and \textit{$k$-nearest} on the existing SOTA, CFEAR. Then, we illuminate the outcomes yielded by our comprehensive algorithm.
Taking into account the characteristics of each dataset to be described in section \ref{sec:k-near}, we present the odometry results with variable $k$=10 for the Ulsan sequence and $k$=50 for the Pohang sequence.
The application of translational/rotational \ac{APE} has been utilized to gauge the trajectory outcomes, the highlighted in bold represent the optimal results. The subsequent section \ref{sec:ablation} provides a meticulous examination of our results, which differ based on varying configurations.

\begin{table*}[t!]
\resizebox{\linewidth}{!}{
\begin{tabular}{llcccccccc}
\textbf{Method} & \textbf{Class(LiDAR/Radar)} & \textbf{Ulsan 01} & \textbf{Ulsan 02} & \textbf{Ulsan 03} & \textbf{Pohang00-b}& \textbf{Pohang01-b} & \textbf{Pohang00-c} & \textbf{Pohang00-d} & \textbf{Pohang01-d} \\ \hline\hline
DLO\cite{chen2022direct} & Dense(L) & 438.14/73.36   & 418.99/14.19 & 745.16/111.06 & 35.61/2.97 & \textbf{1.11}/0.61 & - & -  & - \\ 
Kiss-ICP\cite{vizzo2023kiss}& Dense(L) & 201.64/107.18  & 397.85/45.73 & 533.91/110.96 & 19.98/11.39 & 9.58/5.02 & - & -  & -\\  \hline
%PhaRaO\cite{park2020pharao} & Dense(R) & -/-  & -/- & -/- & -/- & -/- & -/- & -/-  & -/- \\ 
HERO\cite{burnett_rss21} & Semi-Sparse(R) & 223.54/95.15  & 147.98/84.71 & 532.67/94.72 & 53.12/36.35 & 67.11/49.46 & 77.20/32.56 & 275.86/76.29  & 234.67/79.06 \\ 
CFEAR\cite{adolfsson2021cfear} & Sparse(R)  & 238.75/87.71  & 419.25/24.56 & 516.01/78.55 & \textbf{9.51}/1.37 & 9.71/1.16 & 4.46/2.76  & 631.00/48.38  & 197.86/41.88 \\  \hline
Contour & Sparse(R)  & 58.32/105.15 & 19.14/3.93 & 590.08/96.50 & 11.04/0.77 & 5.98/1.30 & 6.37/\textbf{0.71}  & 758.73/81.38 & 94.27/8.41 \\
%CFEAR (KOC) & Sparse & 185.11/0.53   & 36.1/0.09 & 76.36/0.27& 457.73/0.34& 1.05/0.34 & 1.05/0.34 & 1.05/0.34 & 1.05/0.34 & 1.05/0.34 \\ 
%LodeStar (LK) & Semi-Sparse & 29.39/0.14 & 20.49/0.09& \textbf{23.29}/0.06& 121.11/0.34  & 1.05/0.34  & 1.05/0.34 & 1.05/0.34 & 1.05/0.34 & 1.05/0.34 \\ 
%$k$-nearest(20) & Sparse(R)  & 27.52/7.64 & \textbf{10.82}/7.81 & 177.84/11.50 & 48.75/18.79 & 22.28/7.0 & 9.42/1.86  & 97.83/12.10 & 89.59/12.35 \\
$k$-nearest & Sparse(R)  & 22.93/\textbf{5.24} & 15.20/\textbf{3.31} & 216.85/16.98 & 38.88/14.95 & 14.31/4.83 & 4.40/1.81  & 140.06/20.02 & 52.50/14.67 \\
Proposed & Semi-Sparse(R) & \textbf{14.85}/7.48  & \textbf{14.37}/3.59 & \textbf{25.99}/\textbf{3.38} & 10.19/\textbf{0.65} & 9.05/\textbf{0.48} & \textbf{4.38}/2.73  & \textbf{23.24}/\textbf{3.66}   & \textbf{29.48}/\textbf{3.46} \\  \hline
\end{tabular}}
\caption{Translational and Rotational Absolute Pose Error for Maritime Radar and LiDAR Odometry (Trans(m)/Rot(deg))}\label{tab:result}
\vspace{-2mm}
\end{table*}

\subsection{Qualitative Performance}
\begin{figure*}[!t]
    \centering
    \begin{minipage}{0.44\textwidth}
    \subfloat[\texttt{Ulsan03}\label{fig:ablaa}]{
	\includegraphics[width=0.85\columnwidth]{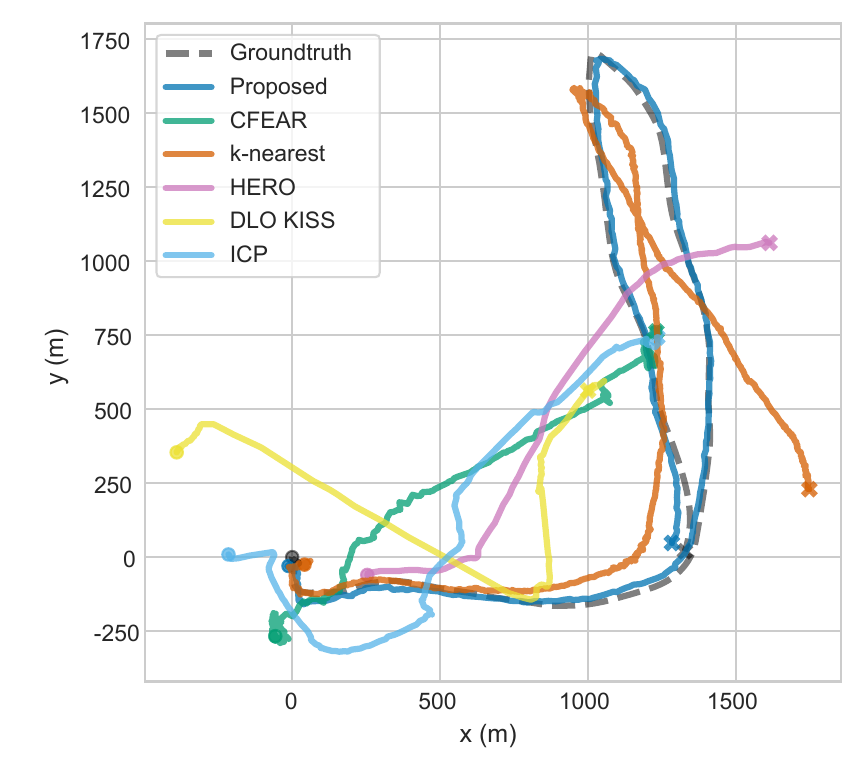}
    }\\
    \subfloat[\texttt{Pohang01-d}\label{fig:ablab}]{
		\includegraphics[clip, trim=0cm 1.5cm 0cm 2.5cm,width=0.85\columnwidth]{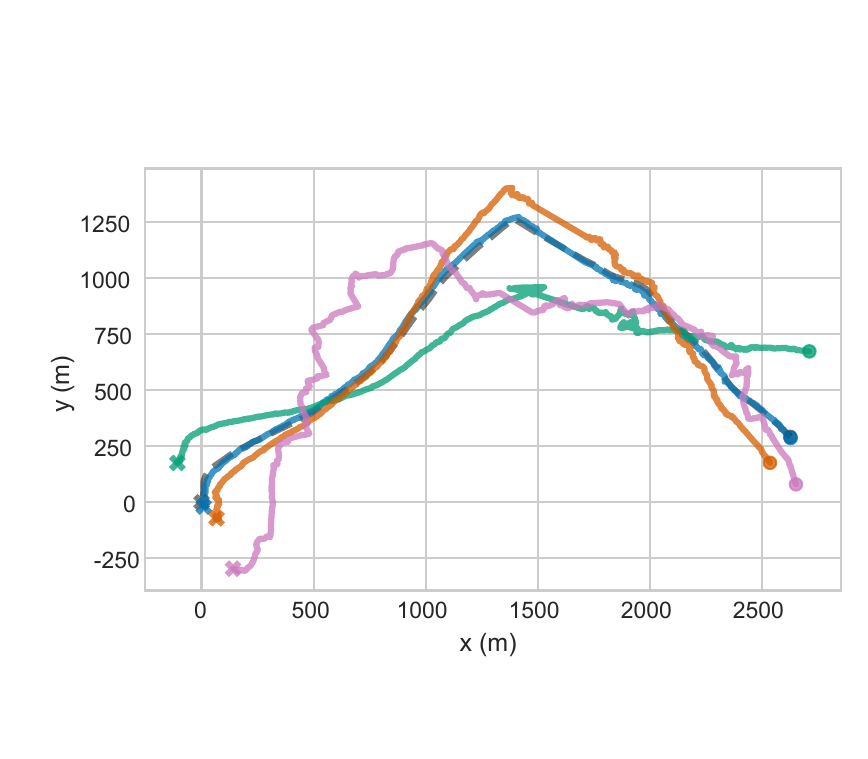}
    }
    \end{minipage}
    \hspace{-6mm}
    \begin{minipage}{0.55\textwidth}
    \subfloat[Pointcloud map based on odometry in Ulsan Sequence\label{fig:ablac}]{
    \includegraphics[width=\columnwidth]{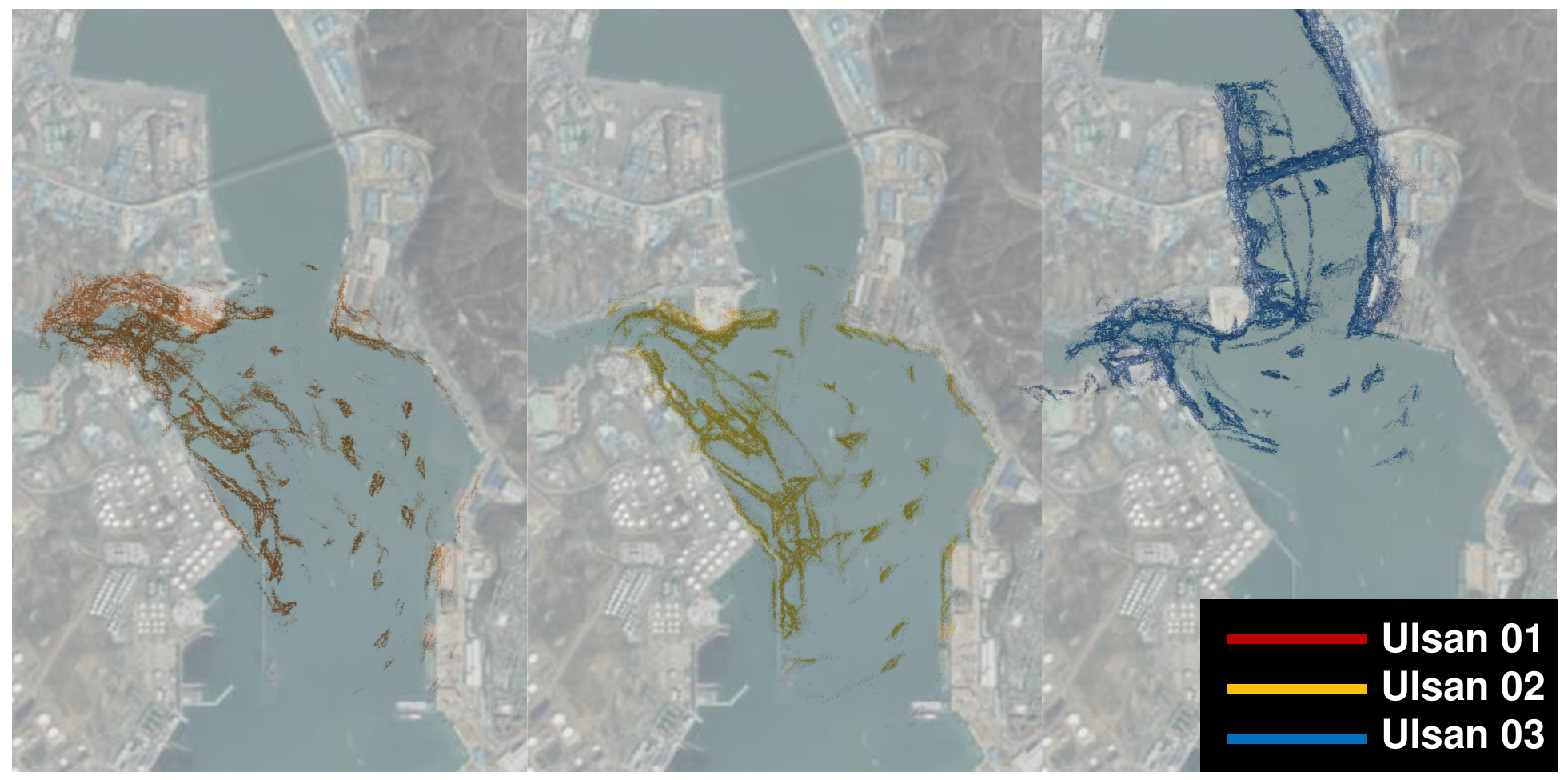}
    }\\
    \subfloat[\texttt{Pohang00-b, c, d} (Rotation)\label{fig:ablad}]{
	\includegraphics[width=0.5\columnwidth]{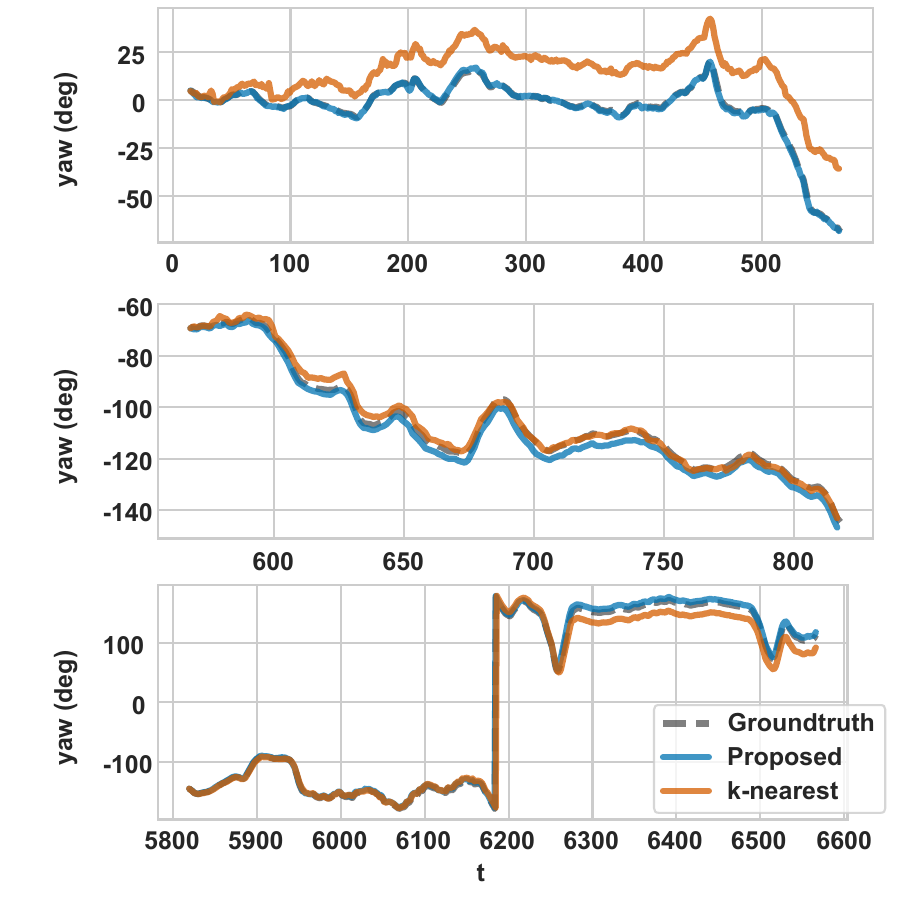}
    }
    \subfloat[\texttt{Ulsan01, 02, 03} (Rotation)\label{fig:ablae}]{
	\includegraphics[width=0.5\columnwidth]{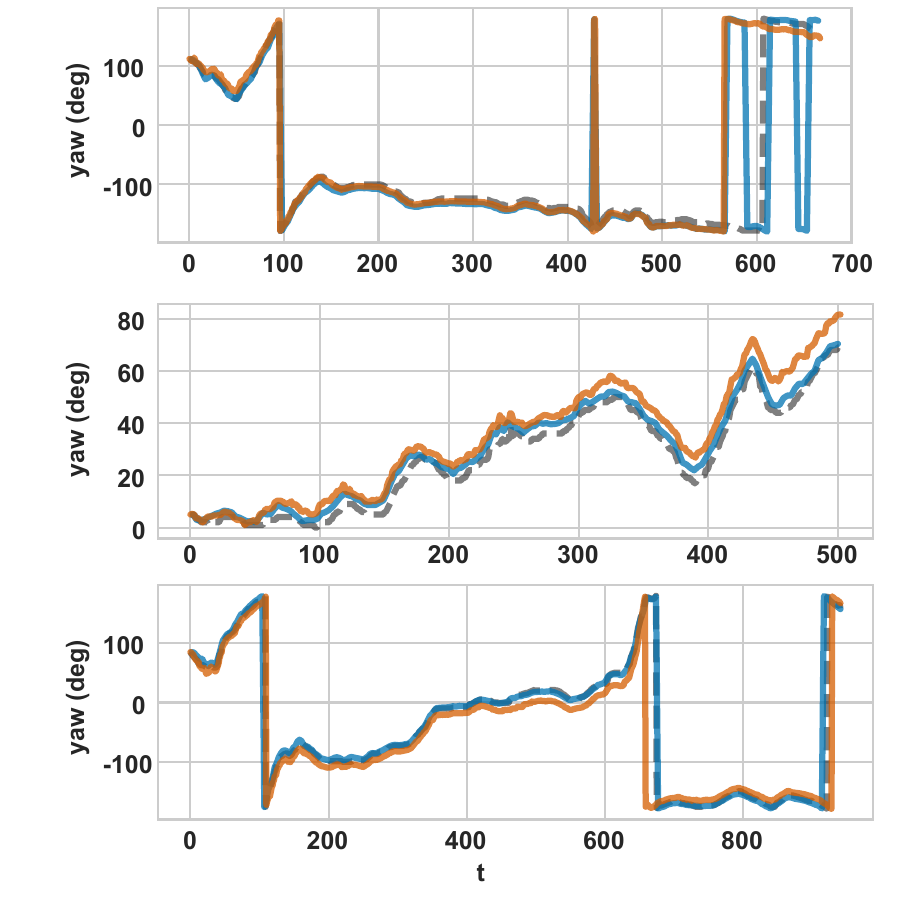}
    }
    \end{minipage}
    \caption{(a) and (b) are the evaluation of odometry estimation techniques in maritime environments. General odometry estimation techniques often fail to generate an accurate trajectory. When the proposed marine feature was incorporated into CFEAR, the system demonstrated improved, albeit approximate, tracking capabilities. Notably, the inclusion of the LodeStar descriptor was able to provide a precise solution for maritime odometry. A detailed comparative analysis is presented in (d) and (e), results before and after the deployment of the descriptor.}
    \label{fig:abla}
    \vspace{-3mm}
\end{figure*}

\figref{fig:abla}\subref{fig:ablaa} and \ref{fig:abla}\subref{fig:ablab} present trajectory estimation results for multiple odometry estimation techniques applied to \texttt{Ulsan03} and \texttt{Pohang01-d}, representing a challenging route. Due to the ineffectiveness of the LiDAR data in \texttt{Pohang01-d}, only the \texttt{Ulsan03} was utilized for LiDAR-based methods. The figure reveals that while robust and reliable techniques for LiDAR and Radar may be valuable in on-ground contexts, they are not directly transferable to maritime radar applications. Only the trajectories obtained from the marine feature and descriptor-aided methods demonstrate an approximate course tendency aligned with the ground truth.
The discrepancies observed in other methods can be attributed to the inherent challenges of marine navigation, specifically the steep turns and scarcity of consistent features.
To provide a focused examination of the proposed method's efficacy without disrupters, trajectories were plotted exclusively for the marine feature-only method and the full proposed algorithm to yield the final results.

\subsubsection{Ulsan Sequence}
%The odometry estimation outcomes, derived from an assortment of methodologies for the Ulsan sequence, are illustrated in \figref{fig:ulsan1}, \ref{fig:ulsan2}, and \ref{fig:ulsan3}. Corresponding numerical values representing the absolute position errors are documented in \tabref{tab:result}. 
\revision{Our algorithm produces a significant enhancement in odometry estimation by exploiting the rotation compensation.} Incorporating the Marine feature into CFEAR yielded a moderate improvement for \texttt{Ulsan01} and \texttt{Ulsan02}. Nonetheless, addressing the sharp rotation in \texttt{Ulsan03} solely with the marine features proved a formidable challenge. The employment of our maritime place descriptor efficaciously rectified the rotation error, subsequently leading to a decrease in translational error. The pointcloud map, constructed using the estimated odometry, is illustrated in \figref{fig:ablac}. This representation accurately captures the surrounding maritime environment.

\subsubsection{Pohang Sequence}
\begin{figure*}[!t]
    \centering
    %\subfloat{
	%\includegraphics[height=0.55\columnwidth]{./figs/radarlidar2.png}
    %}
    %\subfloat{
	%	\includegraphics[height=0.55\columnwidth]{./figs/lidarradar_b.png}
    %}
    %\subfloat{
	%	\includegraphics[height=0.55\columnwidth]{./figs/pohang_radars.png}
    %}
    \includegraphics[width=0.95\linewidth]{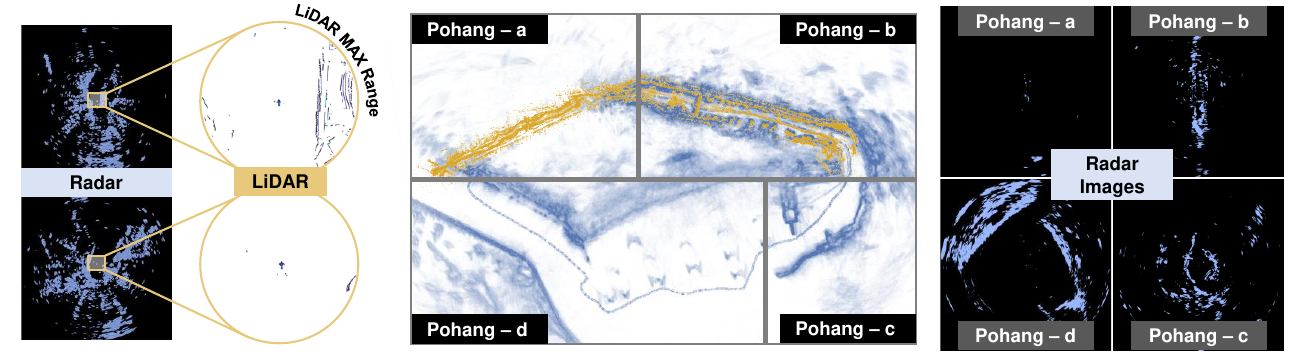}
    \caption{The point cloud mapping outcomes for radar (blue) and LiDAR (yellow) in the Pohang sequence are presented in the center. As evidenced in the left figure, the radar provides a rich set of points, while the LiDAR offers limited information in maritime settings. This paucity of LiDAR detections leads to mapping inaccuracies in expansive maritime regions. Conversely, radar data is sparse in narrow canal regions such as Pohang-a, making it challenging to generate reliable odometry solely using radar.}
    \label{fig:lidarradar}
    \vspace{-3mm}
\end{figure*}

The outcomes from the Pohang Canal Dataset are illustrated in \figref{fig:lidarradar} and also enumerated in \tabref{tab:result}. 
The radar images within the Pohang sequence exhibit complete rotation, devoid of any overlapping regions. As such, implementing our overlap data elimination procedure was deemed unnecessary for this dataset. Our approach for the Pohang data was streamlined, relying solely on contour extraction, $k$-nearest points, and \textit{LodeStar} descriptor.
Since our descriptor predominantly ameliorates rotational errors, its impact on linear areas, such as \texttt{Pohang-b}, and shorter passages like \texttt{Pohang-c}, is relatively subdued. In contrast, a pronounced improvement is evident within the more challenging region of \texttt{Pohang-d}. The near-coastal area, being inherently sparse in features, poses significant challenges for both detection and tracking. Nevertheless, our methodology adeptly navigates these extreme conditions, leading to a recalibrated and refined odometry estimation.

\subsection{Performance of Individual Factors}
\label{sec:ablation}
\begin{figure}[!t]
\vspace{-4mm}
    \centering
%    \subfloat[$k$-nearest]{
%	\includegraphics[width=0.48\columnwidth]{./figs/lk_comp.png}
%    }
%    \subfloat[Contour + $k$-nearest]{
%	\includegraphics[width=0.48\columnwidth]{./figs/lck_comp.png}
%    }
    \includegraphics[width=0.88\columnwidth]{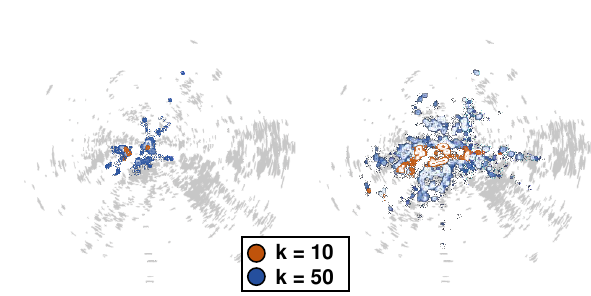}
    \caption{The left depicts the k-nearest points without contour extraction, whereas the right illustrates the post-contour extraction. For the significant values of k, both cases yield precise point normal values. As the number of points reduces, deriving accurate point normals becomes challenging, leading to potential matching discrepancies. However, the point cloud augmented with contour details facilitates the computation of point normals even with a reduced point count.}
    \label{fig:kck_comp}
    \vspace{-5mm}
\end{figure}

For the generality of the algorithm, \revision{the result shown in} \tabref{tab:result} utilized all the marine feature components and Lodestar descriptor. However, the feature extraction method exhibits potential dataset dependency.
In \tabref{tab:dense_abla} and \ref{tab:sparse_abla}, we illustrated the distinct implications of each phase within our algorithmic approach. An exploration into the interrelationships among the feature factors was demonstrated in \figref{fig:kck_comp}.
%Our marine feature extraction can be partitioned into three primary components: contour delineation, $k$-nearest point extraction, and overlap elimination. In the context of dense datasets, the accuracy of k-nearest neighbor algorithms is adversely affected by noise ratio. Conversely, when applying contour extraction techniques to sparse datasets, there's a potential risk of eliminating all data points that constitute the contour context. For linear datasets, the process of overlap elimination becomes redundant.
We conducted an in-depth comparison across the outcomes derived from each type of the datasets, subsequently integrating our distinct descriptor into every algorithmic variant. 
%We corroborated the efficacy and impact of marine features and the descriptor by analyzing comprehensive results. 

\subsubsection{Contour Extraction in Dense Dataset}
Contour extraction gains significance when noise emanates from the coastal surface area. Yet, steep curvature and point cloud sparsity can lead to erroneous matches, as evidenced by \texttt{Ulsan03} and \texttt{Pohang-d} in \tabref{tab:result}. Our validation, as presented in \tabref{tab:dense_abla}, underscores the efficacy of contour extraction in dense datasets with linear characteristics. While the contour extraction in sparse datasets might lead to omitting valuable features, it can markedly augment robustness when generating accurate coastal contours.

\begin{table}[!t]
    \centering
    \resizebox{\columnwidth}{!}{
        \centering
        \begin{tabular}{l|ccc}
        \textbf{Linear \& Dense} & \textbf{\texttt{Ulsan02}} & \textbf{\texttt{Pohang00-b}} & \textbf{\texttt{Pohang01-b}} \\ \hline\hline
        %CFEAR      & 419.24/24.56  & 9.51/1.37 & 9.71/1.16 \\ 
        Contour  & 19.14/3.93 & 11.04/0.77  & 5.98/1.30\\ \hline
        Contour + Overlap & 34.40/3.51 & -  & -\\ 
        Contour + $k$-nearest  & 12.62/6.32 & 9.32/0.83 & 9.26/1.46  \\ 
        Contour + LodeStar  & 12.60/3.16 & 11.57/0.85  & 9.57/0.92\\ 
        Contour + Overlap + LodeStar & 11.85/2.36 & -  & -\\  
        Contour + $k$-nearest + LodeStar  & 11.41/3.55  & 10.19/0.65 & 9.05/0.48 \\\hline
        %Marine Feature  & 15.20/3.31  & 48.75/18.79 & 22.28/7.05 \\ 
        %Marine Feature (O) & 14.91/3.74  & - & - \\ 
        %Marine Feature + LodeStar  & 15.23/3.92  & 36.44/9.78 & 24.07/10.39 \\
        %Marine Feature (O) + LodeStar   & 16.87/4.39 & - & - \\\hline
        \end{tabular}}
    \caption{Effects of Contour in Linear and Dense Dataset}\label{tab:dense_abla}
    \vspace{-3mm}
\end{table}

\begin{table}[!t]
    \centering
    \resizebox{\columnwidth}{!}{
        \centering
        \begin{tabular}{l|ccc}
        \textbf{Curved \& Sparse} & \textbf{\texttt{Ulsan03}} & \textbf{\texttt{Pohang00-d}} & \textbf{\texttt{Pohang01-d}} \\ \hline\hline
        %CFEAR      & 516.01/78.55  & 631.00/48.38 & 197.86/41.88 \\ \hline
        %Contour  & 590.08/96.50 & 758.73/81.38  & 94.27/8.41\\ 
        %Contour + Marine Feature  & 102.56/22.41 & 788.98/95.83 & 165.73/20.97  \\ 
        %Contour + Marine Feature + LodeStar  & 40.58/3.99  & 23.24/3.66 & 29.48/3.46 \\\hline
        $k$-nearest  & 216.85/16.98  & 140.06/20.02 & 52.50/14.67 \\ \hline
        $k$-nearest + Overlap & 133.56/12.28  & - & - \\ 
        $k$-nearest + LodeStar  & 49.06/3.61  & 24.30/3.47 & 8.69/1.30 \\
        $k$-nearest + Overlap + LodeStar   & 21.88/1.96 & - & - \\\hline
        \end{tabular}}
    \caption{$k$-nearest Feature in Curved and Sparse Dataset}\label{tab:sparse_abla}
    \vspace{-7mm}
\end{table}

\subsubsection{$k$-nearest Candidates in Sparse Dataset}\label{sec:k-near}
%In the exploration of the $k$-nearest point candidates, it is the most important feature factor conserving the submap of the place. 
Across all datasets, the incorporation of $k$-nearest candidates markedly improved accuracy as depicted in \tabref{tab:result}. Moreover, \tabref{tab:dense_abla} and \ref{tab:sparse_abla} demonstrate a complementary relationship with \textit{LodeStar} descriptor. Particularly in sparse datasets, $k$-nearest points preserve the scene's context, facilitating the tracking of approximate trajectories.

Additionally, we evaluated the outcomes resulting from modifications in the quantity of nearest points. Considering that the radar image possesses a resolution of $2.71m/pixel$, APE values are nearly uniform across generic instances. However, the results for the inner port zone within the Pohang sequences (\texttt{Pohang00-b}, \texttt{Pohang01-b}) illustrating a discernible decline in accuracy as the number of points amplifies. Such findings underscore the necessity for a sufficient quantity of nearest points in regions characterized by intricate and a strong radar signals. Contrarily, \texttt{Ulsan03} yielded superior outcomes with a diminished point numbers. These observations substantiate the premise that the exclusion of certain points can be beneficial in complex but expansive areas. %In scenarios encompassing both intricate and expansive regions (\texttt{Pohang00-d}, \texttt{Pohang01-d}), it becomes evident that focusing on both ends surpasses a nebulous configuration. 
\revision{To clarify, a larger value of $k$ is beneficial in areas with complex environments and strong radar signals, whereas a smaller value is appropriate in regions that are sparse and have weak radar signals.}
Details are written in \tabref{tab:k-near}, and bolded figures support these results.

\begin{table}[t!]
\resizebox{\linewidth}{!}{
\begin{tabular}{cc|cccc}
%\multicolumn{2}{c}{\multirow{2}{*}{\textbf{Dataset}}} & \multicolumn{4}{c}{k-nearest}  \\
\multicolumn{2}{c}{\textbf{Dataset}}                        & k=10       & k=20     & k=50 & k=100      \\ \hline\hline
\multicolumn{1}{c}{\multirow{3}{*}{Ulsan}} & \texttt{01}       & 14.85/7.48    & 20.24/12.42 & 24.27/15.99 & 21.11/7.68 \\ 
\multicolumn{1}{c}{}                        & \texttt{02}    & 14.37/3.59    & 19.59/6.92 & 15.64/5.23  & 8.16/1.92             \\ 
\multicolumn{1}{c}{}                        & \texttt{03} & 25.99/3.38   & 29.37/3.08 & 40.21/3.74 & \textbf{60.41/7.49}  \\ \hline
\multicolumn{1}{c}{\multirow{5}{*}{Pohang}} & \texttt{00-b}  & \textbf{68.40/12.72}  & 18.02/2.19  & 10.19/0.65 & 11.97/2.37  \\
\multicolumn{1}{c}{}                       & \texttt{01-b}  & \textbf{34.60/9.80}  & 11.79/1.77 & 9.05/0.48  & 6.66/0.52  \\
\multicolumn{1}{c}{}                       & \texttt{00-c}  & 7.77/2.56  & 6.62/2.73 & 4.38/2.73  & 4.67/1.06  \\
\multicolumn{1}{c}{}                       & \texttt{00-d}  & 14.82/2.10  & 17.52/1.28 & 23.24/3.66  & 13.19/3.22  \\
%\multicolumn{1}{c}{}                       & 1-3  & 102.83/3.12  & 104.07/\textbf{3.12} & 101.63/3.12  & 103.81/\textit{3.12}  \\
\multicolumn{1}{c}{}                       & \texttt{01-d}  & 24.25/2.20  & 30.79/2.42 & 29.48/3.46  & 22.35/1.93  \\\hline
\end{tabular}}
\caption{Results for Modifications in the $k$-nearest Points}\label{tab:k-near}
\vspace{-3mm}
\end{table}

\begin{figure}[!t]
    \vspace{-1mm}
    \centering
	\includegraphics[width=0.85\columnwidth]{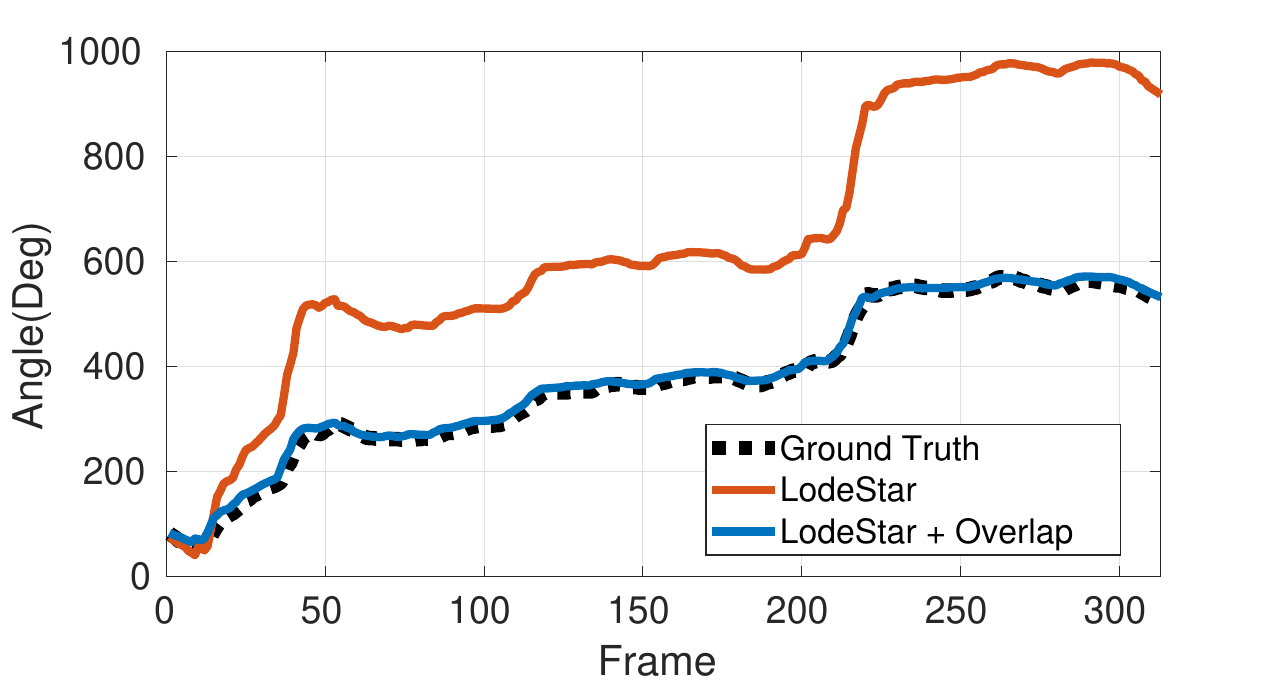}
    \caption{The rotation estimation only with the \textit{LodeStar} in the Ulsan03 sequence, highlighting significant rotational drifts near frames 40 and 210. In the absence of overlap elimination, rotation estimation is vulnerable, and accumulated angles demonstrate substantial errors. However, with the integration of overlap elimination, producing precise angle estimation results becomes feasible.}
    \label{fig:lodeandover}
    \vspace{-5mm}
\end{figure}

\subsubsection{Overlap Elimination in Curved Dataset}
Overlap elimination mitigates the impact of repeated scans, particularly those arising from steep rotational movements. In linear trajectories, the implications of overlap elimination are relatively inconsequential. Yet, as illustrated in \tabref{tab:sparse_abla}, this refinement can modestly reduce error rates. Notably, its integration with \textit{LodeStar} descriptor results in significant performance enhancements. This particular behavior is evident in \figref{fig:lodeandover}. Without the implementation of overlap elimination, the estimation experiences significant error drift, particularly in areas of rotation. However, following the removal of overlapping regions, it becomes feasible to achieve near ground-truth level accuracy in estimating the vessel's rotation, even when relying solely on \textit{LodeStar}-based estimations.

\subsubsection{LodeStar Descriptor Performance}
The results are presented both prior to and subsequent to the application of the descriptor for each feature methodology in \tabref{tab:dense_abla} and \ref{tab:sparse_abla}. As illustrated in \figref{fig:abla}, the vessel's acute maneuvering significantly impacts odometry tracking. Nevertheless, our descriptor effectively addresses all instances of rotational discrepancies. The diminutive rotation angle variance, such as linear or minimal curvature routes, renders the descriptor's function inconsequential. The intricacy of the trajectory directly influences the outcomes, with increasing complexity bolstering the efficacy of our descriptor. As delineated in \figref{fig:lodeandover}, we have validated the robustness of our descriptor under challenging environmental conditions.

\section{Conclusion}
\label{sec:conclusion}
This paper introduced a novel maritime radar descriptor that significantly diminishes odometry errors. Our investigation into the marine environment context utilized three distinct feature extraction methodologies, further aiding in the enhancement of the odometry outcomes. Particularly for sharp turning routes, our descriptor is indispensable for accurately tracking and predicting the vessel's trajectory. In spite of the advancements made, minor challenges remain. The absence of a pointcloud results in unsuccessful matching and tracking. Conversely, a reduced quantity of $k$-nearest points augments accuracy during acute turning. Determining the optimal parameter or applying the methods such as contour extraction is imperative.
As prospects, we aim to integrate our odometry estimation approach with place recognition techniques to implement a \ac{SLAM} using marine radar.

% \newpage
% \newpage

%\section*{ACKNOWLEDGMENT}
\balance
\small
\bibliographystyle{IEEEtranN} %citeauthor
\bibliography{string-short,references}

\end{document}